\documentclass[conference]{IEEEtran}
\IEEEoverridecommandlockouts
\usepackage{cite}
\usepackage{amsmath,amssymb,amsfonts}
\usepackage{algorithmic}
\usepackage{graphicx}
\usepackage{textcomp}
\usepackage{xcolor}
\usepackage{algorithm}
\usepackage{booktabs}
\def\BibTeX{{\rm B\kern-.05em{\sc i\kern-.025em b}\kern-.08em
    T\kern-.1667em\lower.7ex\hbox{E}\kern-.125emX}}
\begin{document}

\title{Graph Attention Network for Predicting Duration of Large-Scale Power Outages Induced by Natural Disasters\\

}

\author{\IEEEauthorblockN{Chenghao Duan}
\IEEEauthorblockA{\textit{School of Electrical and Computer Engineering} \\
\textit{Georgia Institute of Technology}\\
Atlanta, GA, USA \\
cduan8@gatech.edu}
\and
\IEEEauthorblockN{Chuanyi Ji}
\IEEEauthorblockA{\textit{School of Electrical and Computer Engineering} \\
\textit{Georgia Institute of Technology}\\
Atlanta, GA, USA \\
jichuanyi@gatech.edu}

}

\maketitle

\begingroup
\renewcommand\thefootnote{}
\footnotetext{© 2025 IEEE. Personal use of this material is permitted. Permission from IEEE must be obtained for all other uses, in any current or future media, including reprinting/republishing this material for advertising or promotional purposes, creating new collective works, for resale or redistribution to servers or lists, or reuse of any copyrighted component of this work in other works. Accepted for publication in: Proceedings of the 2026 IEEE PES International Meeting, 18 – 21 January 2026, Hong Kong SAR, China}
\addtocounter{footnote}{-1}
\endgroup

\begin{abstract}

Natural disasters such as hurricanes, wildfires, and winter storms have induced large-scale power outages in the U.S., resulting in tremendous economic and societal impacts. Accurately predicting power outage recovery and impact is key to resilience of power grid. Recent advances in machine learning offer viable frameworks for estimating power outage duration from geospatial and weather data. However, three major challenges are inherent to the task in a real world setting: spatial dependency of the data, spatial heterogeneity of the impact, and moderate event data. We propose a novel approach to estimate the duration of severe weather-induced power outages through Graph Attention Networks (GAT). Our network uses a simple structure from unsupervised pre-training, followed by semi-supervised learning. We use field data from four major hurricanes affecting $501$ counties in eight Southeastern U.S. states. The model exhibits an excellent performance ($>93\%$ accuracy) and outperforms the existing methods XGBoost, Random Forest, GCN and simple GAT by $2\% - 15\%$ in both the overall performance and class-wise accuracy. 

\end{abstract}


\section{Introduction}\label{Intro}

Natural disasters such as hurricanes, wildfires, and winter storms have caused large-scale power outages in the U.S., leaving millions without electricity for days and resulting in annual economic losses of \$150 billion \cite{carvallo2024weather, DOE2018report}. These outages disproportionately impact disadvantaged communities \cite{doe2021j40, ganz2023socioeconomic}. Thus, enhancing energy grid and community resilience has been widely advocated by researchers, industry, and government \cite{xu2024resilience, excutiveorder2013}, with rapid recovery as a critical component \cite{doe2021climate, white2021j40}. Reliable prediction of the recovery speed would enable efficient emergency response, resource allocation, and infrastructure enhancement\cite{ stankovic2022methods, afsharinejad2021large}. 

The research problem we study is to estimate outage duration given the weather conditions of a severe weather event, geospatial information on outage severity, and demographic characteristics of affected communities using machine learning approach, i.e., to learn a mapping between outage duration and the given variables. The learning problem exhibits three unique challenges: (a) Spatial dependence: Weather-induced power outages are often spatially correlated; (b) Spatial heterogeneity: The severity of weather events varies spatially, resulting in heterogeneous recovery and resilience; (c) Moderate event data: Number of historical extreme hurricanes is limited.

To address the challenges, we develop a Graph Attention Network (GAT) \cite{velivckovic2017graph} to learn a mapping between outage duration and spatial feature variables. The model we develop aims to characterize, through a simple structure, the spatial dependence of power outages induced by hurricanes. At a pre-training stage, the structure of the embedding layer is obtained through clustering weather conditions and outage characteristics to address spatial heterogeneity. Such an embedding structure is simple: we choose the simplest case of two clusters for the embedding layer to form a Bimodal Graph Attention Network (BiGAT). Parameters are then learned for the embedding, attention and output layer through semi-supervised learning, where the data are labeled at partial locations. The model (BiGAT) is implemented in a real-world context. Our data set consists of large-scale power outages induced by four major hurricanes from 2017 to 2020 in the southeast states of the U.S. The outage events affected $501$ counties and $11.6$ million customers. 

Prior work on weather-induced power outage recovery has focused on statistical and machine learning methods. Applied models include tree-based models, such as Random Forest and XGBoost \cite{nateghi2014forecasting, johnson2024can}, and neural networks, including deep and convolutional networks \cite{jaech2018real, yu2018deepsolar}. These models do not explicitly capture spatial characteristics and often require large, multi-year daily-operation data \cite{fang2022failure}; with limited data, the spatial scope has to be reduced\cite{nateghi2014forecasting}, which would not be suitable for the scale of extreme weather events. Graph neural networks (GNNs) have been applied to power grids, mainly for power flow, load, and outage/attack detection \cite{liao2021review}, using synthetic or standard test data. Few works use real data for outage detection \cite{owerko2018predicting}, and none address outage duration or impact. The research challenges in Section \ref{Intro} remain open for real-world outage learning from natural disasters.

\section{Method}

\subsection{Variables}
We select the following variables for GAT. The output of the GAT is power outage duration. The outage durations signify the impact and recovery speed of weather-induced large-scale power failures. The feature variables of the GAT consist of the following: (a) Hurricane wind swath as an exogenous cause of power outages; (b) The maximum number of affected customers that represents the failure impact but not recovery; (c) Geospatial characteristics of affected regions including population density and area; (d) Socioeconomic vulnerability of affected communities. 

\subsection{Data}\label{DurThres}
We focus on four major Atlantic hurricanes (2017–2020) in the Southeastern U.S., selected by criteria from prior work \cite{ganz2023socioeconomic}. Data sources include: PowerOutage.com \cite{powerUS22} for county-level, time-varying affected customer counts (with outage duration defined from peak disruption to restoration below 5\% of customers served); National Hurricane Center best track data (HURDAT2) \cite{HURDAT2} for wind swath (34–49, 50–63, 64+ knots); U.S. Census data \cite{USCensusBureau2020} for county population and area; and Centers for Disease Control SVI \cite{flanagan2018measuring} for socioeconomic vulnerability. Outage durations are categorized as short ($<$2 days), medium (2-6 days), or long ($>$6 days), aligned with U.S. Department of Energy restoration tiers \cite{doe2022playbook}.

\subsection{Spatial correlation and heterogeneity}\label{geoclustering}
We explore spatial correlation in our data as a prerequisite for designing a GAT structure. We measure the spatial correlation of duration and peak outages using Moran's I \cite{moran1950notes}. The results of four hurricanes are summarized as follows:

\begin{table}[h]
\caption{Global Moran's I for each hurricane event}
\label{tbl-morani}
\begin{center}
\begin{small}
\begin{sc}
\begin{tabular}{lcc}
\toprule
Events & Log(DURT) & Log(Peak) \\
\midrule
Florence  & $0.750^{*}$ & $0.583^{*}$ \\
Irma      & $0.755^{*}$ & $0.570^{*}$ \\
Laura     & $0.747^{*}$ & $0.570^{*}$ \\
Michael   & $0.760^{*}$ & $0.422^{*}$ \\
\midrule
\multicolumn{3}{p{0.3\textwidth}}{\small \textit{\tiny $^*$ indicates the rejection of the null hypothesis for a Moran's I value to be 0.}}
\end{tabular}
\end{sc}
\end{small}
\end{center}
\vskip -0.1in
\end{table}

Moran’s I show a positive spatial correlation for all events, particularly on outage durations. This demonstrates that spatial correlation is inherent in hurricane-induced power outages. Such spatially correlated outage data motivates the use of GAT networks. 

\begin{figure}[!h]
    \centering
    \includegraphics[width=0.8\linewidth]{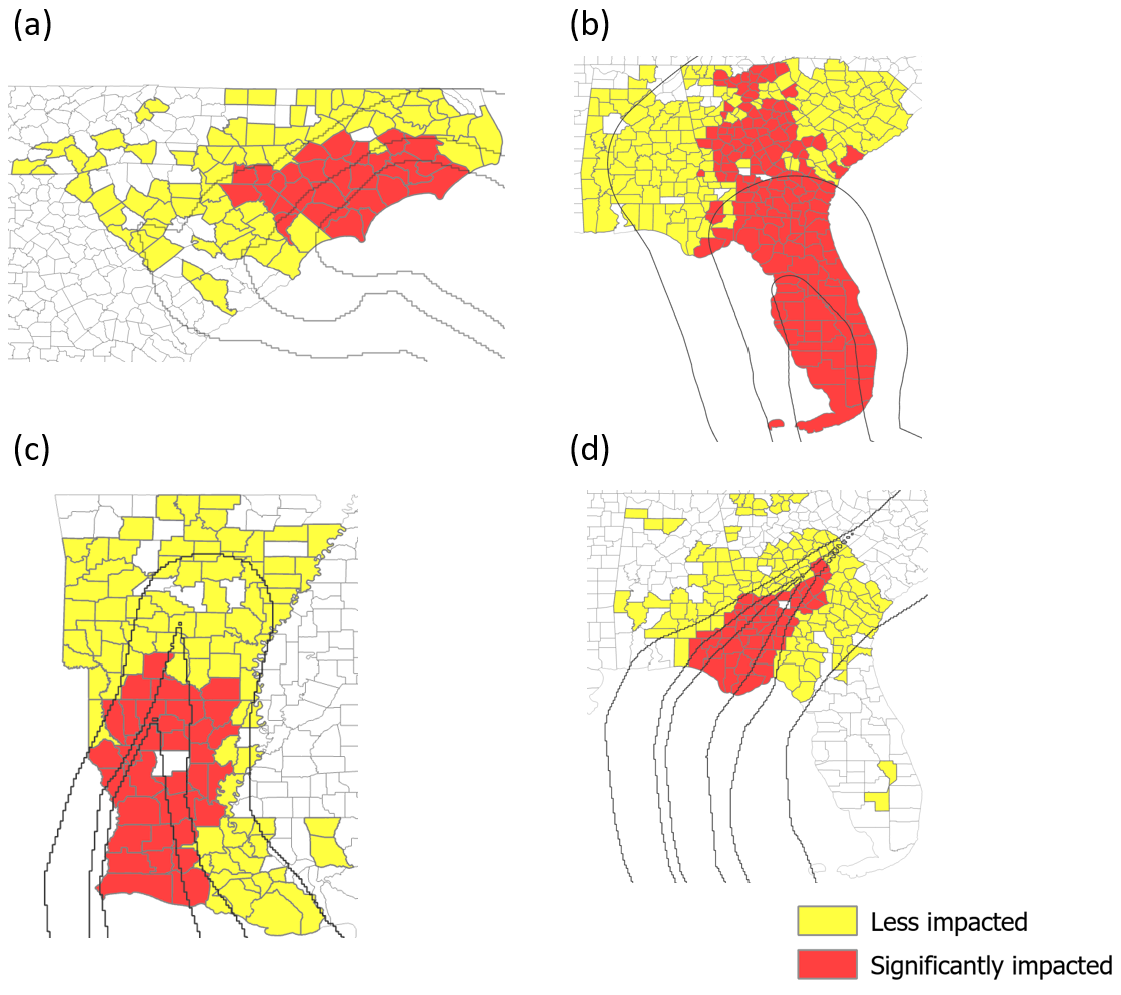}
    \caption{K-means clustering plot for four hurricane events: (a) Florence, (b) Irma, (c) Laura, (d) Michael. $k=2$ is the number of clusters. Red area: significantly impacted counties. Yellow area: less impacted counties. Dark lines: boundaries of wind swaths of 64 knots, 50 knots, and 34 knots counted from inside out.}
    \label{fig:LauraCls}
\end{figure}

We further cluster labeled outage duration and wind swath using the K-means algorithm \cite{lloyd1982least}. These two variables measure the severity of an outage event and weather directly. In Figure \ref{fig:LauraCls}, the affected counties are clustered into two groups. One is clustered around the center of each hurricane track with a higher wind swath and longer outage duration, namely the significantly impacted counties. The other is farther away from the hurricane center, with a lower wind swath and shorter outage duration, namely less impacted counties. These characteristics are reflected by the spatial clusters of the county locations. Hence, the outage data are not only spatially correlated but also spatially heterogeneous, i.e., affected regions are impacted differently at different locations by the hurricanes. This motivates us to construct the structure of the BiGAT to incorporate such spatial characteristics. Note that the cluster number is a model design choice, depending on data and scale of study. We choose $k=2$ for our experiments as the simplest case, given the available data ($\approx 130$ samples per event).

\subsection{Problem formulation}
We formulate the problem of learning a mapping between the feature variables and outage duration using graph-based neural network. Our formulation is based on the following assertions: 
\begin{enumerate}
\item[(a)] Spatially correlated outages are for a given hurricane event (dependence across different hurricanes is not considered in this work).
\item[(b)] The spatial correlation based on the nearest neighbors is the simplest to adopt in this work.
\item[(c)] County-level variables are used because of the available data. The formulation is extensible to other settings.
\end{enumerate}

For a given hurricane event, we let $G = \{V, E, A\}$ denote a graph representation of outages, where $V$ is a set of affected counties (nodes), $E$ consists of spatial adjacent county pairs (edges), and $A$ is a symmetric adjacency matrix. $A_{(i,j)}=1$ if nodes $i$ and $j$ are neighbors, and $A_{(i,j)}=0$, otherwise. Further, we let $X_{v} \in R^d$ represent a feature vector that consists of inputs to node $v$ ($v \in V$). $X=\{X_v\}$ for all $v\in V$ represents values of feature vectors for all nodes. $d=11$ is the input dimension for this work. The 11-dimensional feature vector consists of the maximum number of affected customers, county population, county area, four quantities for social vulnerability, and four for hurricane wind swaths. $Y \in R^{|V| \times 3}$ represents the node-level 3-class duration labels, as defined in Section \ref{DurThres}. 

Our problem can be further posed as node-level classification, i.e., to obtain a GAT model $M$ that predicts the class labels of outage durations $Y$ using node features $X$ and graph structure $G$, 
\begin{align*}
    Y &= M(X, G).
\end{align*}

\subsection{Bimodal GAT}
We develop the BiGAT with a simple GAT structure. The model development includes three aspects: a pre-training stage for node embedding through unsupervised learning, an overall BiGAT network structure, learning and message passing with labeled and unlabeled samples. 

\subsubsection{Pre-training for node embedding} 
In the pre-training stage, we obtain an embedding structure that incorporates spatial correlation and heterogeneity through clustering. As shown in Section \ref{geoclustering}, the data form two clusters as the simplest case we consider. Learnable weights at the node embedding layer are thus assumed to take two sets of distinct values. Specifically, for cluster $k$ and node $v \in V$, we let $\beta^{(k)}$ be the learnable embedding weights for $k=1,2$. The learnable parameter $\beta^{(k)}$ is the same for all nodes in the same cluster. The resulting embedded message for node $v\in V$ is 
\begin{align*}
m_v &= \beta^{(k)} X_{v},
\end{align*}
The pre-training stage determines which (locations of) nodes belong to which cluster but not the values of $\beta^{(k)}$. The latter are to be learned at the training phase. This pre-training stage reduces the model complexity, facilitating learning from the moderate number of hurricane events. For testing (unlabeled) nodes, the cluster labels are inferred by a pre-trained GAT $M_{cluster}$ with the same set of input variables as BiGAT. $M_{cluster}$ is trained with training nodes only.

\begin{align*}
\hat{k}_{test} &= M_{cluster}(X,G).
\end{align*}

\subsubsection{Overall BiGAT structure}
\begin{figure*}[!th]
\centering
\includegraphics[width=0.75\linewidth]{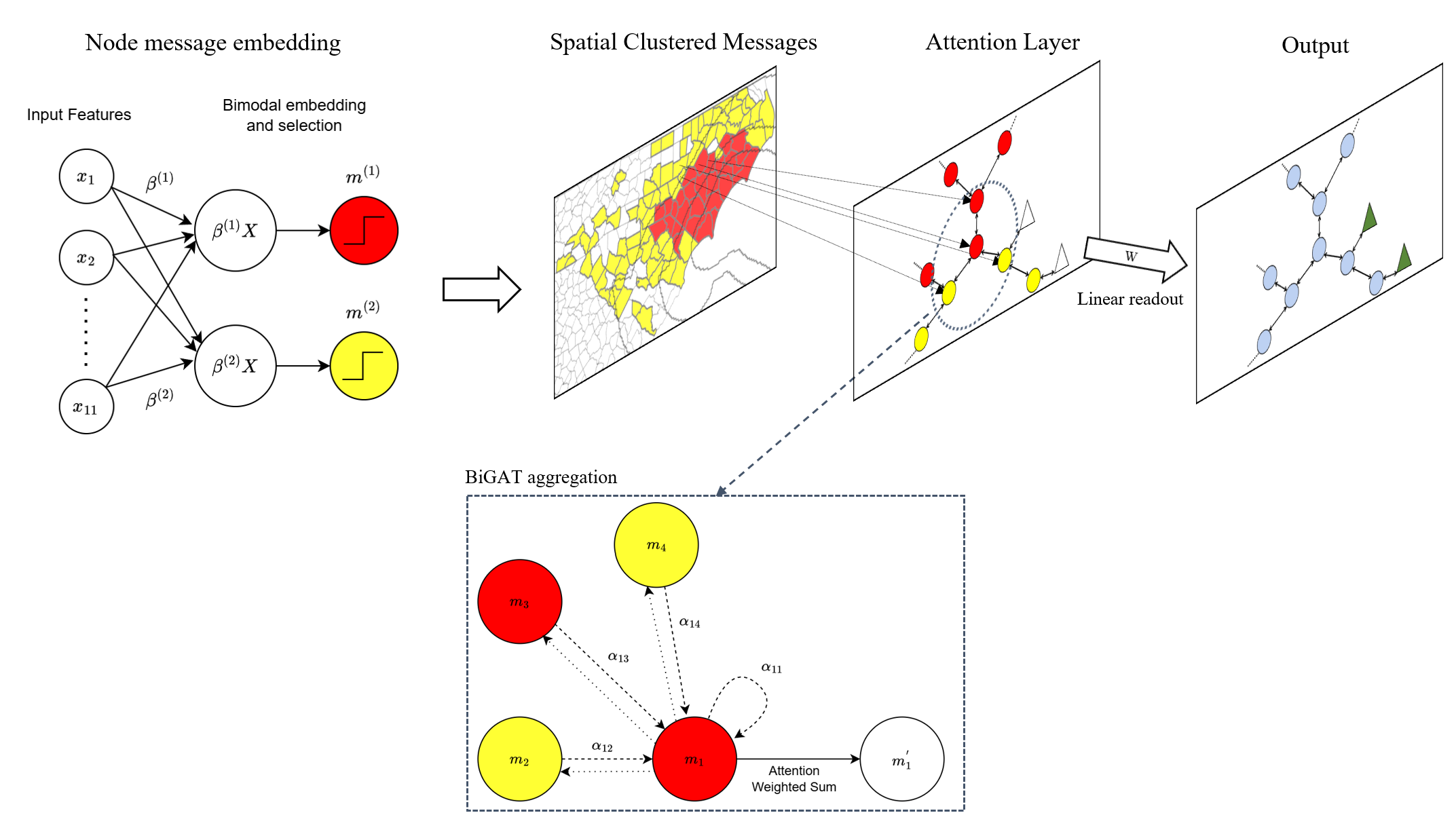}
\caption{Overall BiGAT network. Embedding layer: from input features $X=\{x_1,…x_{11}\}$ to each affected node (county) message $m^{(\cdot)}$. Each node is associated with one message embedding ($\beta^{(1)}$ or $\beta^{(2)}$) based on the cluster label. Structure of the embedding layer is developed from two clusters: Red - severely impacted counties. Yellow - moderately impacted counties. Attention layer: attention weights $\alpha_{ij}$. Output layer: duration labels for two types of counties: Grey circle - labeled counties, Green triangle - unlabeled counties} 
\label{fig:BiGATNet}
\end{figure*}

After the embedding layer, the BiGAT model applies self-attention to each node and aggregates messages from neighbors as in the standard GAT network, as illustrated in Figure \ref{fig:BiGATNet}. For a given pair of nodes $u$ and $v$, the attention weights from node $u$ to neighbor $v$ are calculated as:
\begin{align*}
e_{vu} &= \delta_a(a_{src}^{T} m_{u} + a_{dst}^{T} m_{v}), \\
\alpha_{vu} &= \frac{exp(e_{vu})}{\sum_{i \in N_v}exp(e_{vi})},
\end{align*}
Where $\delta_{a}$ is a LeakyReLU activation function. $N_v$ represents neighbors of $v$. $a_{src}^{T} $ and $ a_{dst}^{T} $ are two learnable parameters for a source- and destination-node respectively. As a special case of BiGAT, the attention coefficients can be further regulated, where $a_{src} = a_{dst}$. This results in direction-agnostic attention coefficients among neighbors, referred to as BiGAT-undirected. Then, the messages are aggregated using the attention weights:
\begin{align*}
    m^{'}_v &= \sum_{u \in N_v}\alpha_{vu} m_u + \alpha_{vv}m_v,
\end{align*}

Finally, a linear readout layer transforms the aggregated messages at each node into outputs, i.e., the categories of outage durations. 
\begin{align*}
    O_v &= Wm_v^{'} + b, \\
    \hat{c_v} &= argmax(O_v).
\end{align*}
The pseudo Algorithm \ref{alg:BiGAT} summarizes the overall pipeline of outage duration classification: 
\begin{algorithm}[!h]
\caption{Overall BiGAT pipeline}\label{alg:BiGAT}
1. Curate graph-structured outage data using geospatial adjacency information.\\
2. Use K-means clustering to determine the structure of the embedding layer. \\
3. Use graph-structured outage data, clustered embedding layer and simplified attention mechanism to train learnable parameters.\\
4. Predict the outage duration classes for testing and evaluate the performance.
\end{algorithm}
\subsection{Experimental setting for training and testing}
The learnable weights of the BiGAT model are trained in a transductive setting\cite{vapnik2006transductive} for each hurricane event. We randomly split the affected counties into training and testing sets, and perform stratified five-fold cross-validation for each hurricane event, with 80\% data for training and the remaining 20\% for testing. The cross-validation is repeated six times with different random splits to obtain robust performance evaluation statistics. This setting aligns with the real-world scenario: During the recovery process, the peak outage numbers are generally available via grid monitoring, and the prediction (testing) is applied in regions where a reliable estimation of outage duration is lacking.

To train the model for node-level multi-class classification tasks, we choose the loss function to be cross-entropy loss and use ADAM optimizer for training. We use full batch training, with a learning rate to be $0.005$ and hidden feature dimension of $3$. The model training and testing are implemented using Pytorch, PyGeometrics, and Sklearn packages on Google Colab platform with default CPU.

\section{Results}
\subsection{Performance metrics}
We employ three metrics to evaluate the performance of the model for each hurricane event: (1) Classification Accuracy; (2) Macro F1 Score; (3) Balanced Accuracy. The classification accuracy evaluates the overall performance of the model by treating all classes equally. The macro F1 score and balanced accuracy provide class-balanced evaluations of performance, taking into consideration of the prior probabilities of the classes. 

\subsection{Model performance}\label{ModelPerformance}
We apply BiGAT to the data sets with the four extreme hurricane-induced power outages. To benchmark the performance, we also apply widely adopted models to the learning of the same data, including XGBoost, Random Forest, single-layered Graph Convolutional Network (GCN) and single-layer GAT. For each benchmark model, we train iteratively over different sets of hyperparameters to ensure the best performance.
 
\begin{figure}[!h]
    \centering
    \includegraphics[width=0.9\linewidth]{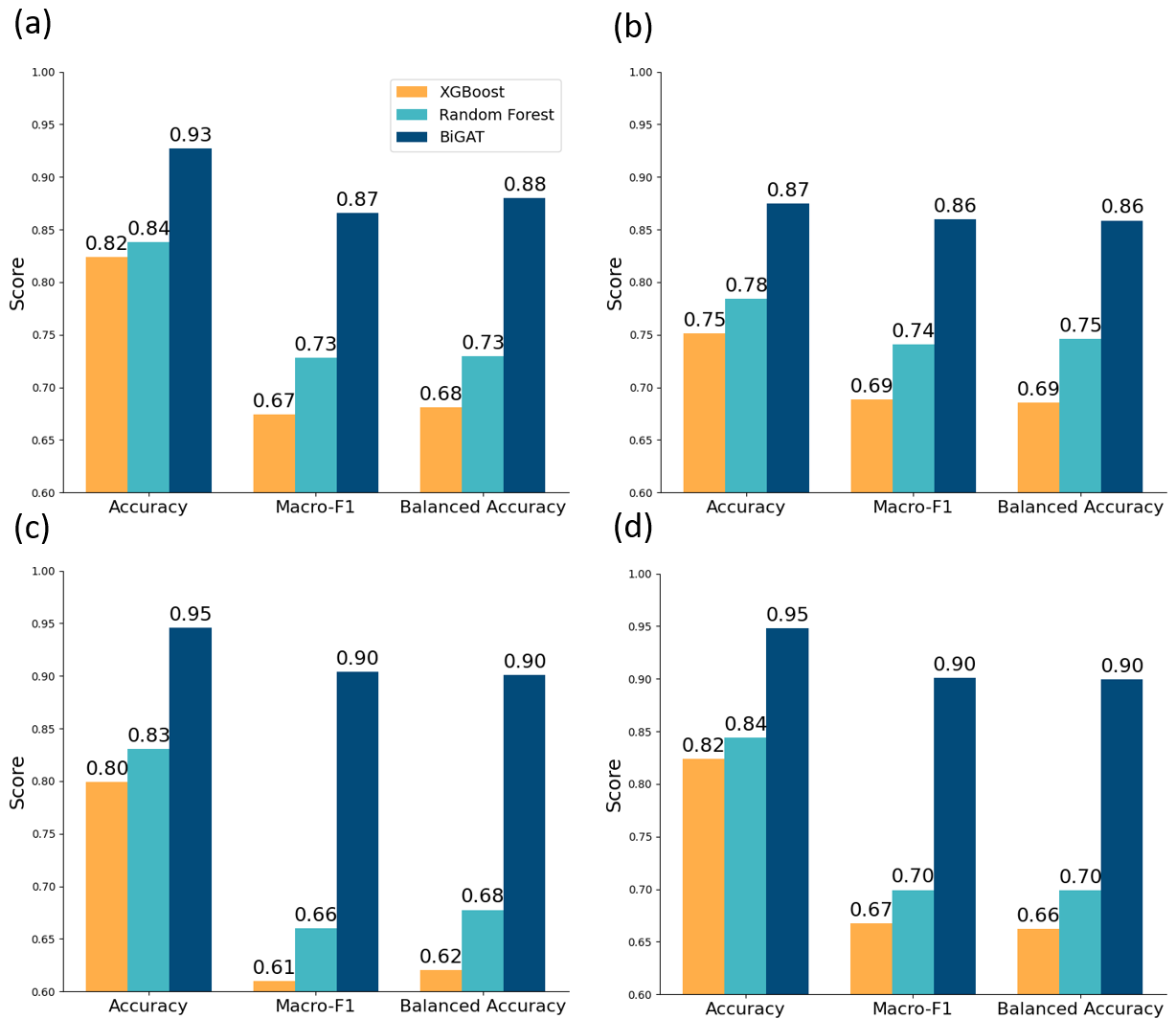}
    \caption{BiGAT versus non-graph models - performance comparative analysis: (a) Florence (b) Irma (c) Laura (d) Michael. Each subplot presents three performance metrics: classification accuracy, macro F1, and balanced accuracy. Models are XGBoost, Random Forest, and BiGAT.}
    \label{fig:PerformanceResultsRF}
\end{figure}

As shown in Figure \ref{fig:PerformanceResultsRF}, BiGAT significantly outperforms XGBoost and Random Forest in all events. Specifically, the classification accuracy improves by $9\% - 15\%$. Macro F1 improves by $0.12 - 0.29$. The balanced accuracy improves by $11\% - 28\%$. The performance enhancement highlights the graph network's strength in capturing spatial dependency in outage data. 

\begin{figure}[!h]
    \centering
    \includegraphics[width=1\linewidth]{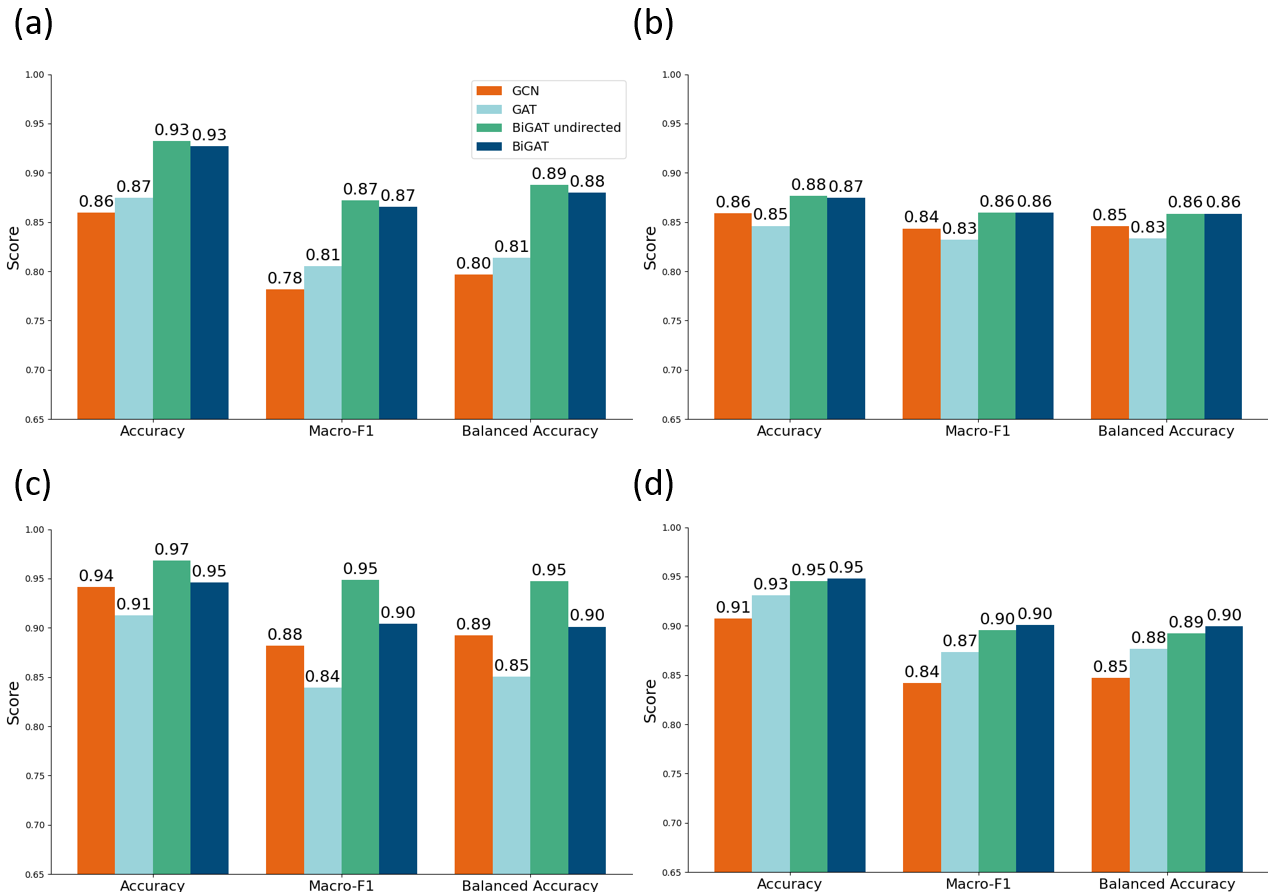}
    \caption{Graph models performance comparative analysis: (a) Florence (b) Irma (c) Laura (d) Michael. Each subplot presents three performance metrics: classification accuracy, macro F1, and balanced accuracy. Models are GCN, GAT, BiGAT undirected, and BiGAT.}
    \label{fig:PerformanceResultsGNN}
\end{figure}

The BiGAT is then compared with single-layered GCN and GAT, as shown in Figure \ref{fig:PerformanceResultsGNN}. BiGAT and BiGAT-undirected both achieve $93\%$ average accuracy over four events, and their macro F1 and balanced accuracy are at least $0.86$. Compared with GCN and GAT, the BiGATs improve the performance metrics by $2\% - 11\%$ for all events. Note that the improvements of macro F1 and balanced accuracy are in general higher than that of the average accuracy. 

We further examine the models' average accuracy on each duration class. The results are shown in Table \ref{ClassAcc}. The results show that the accuracies of medium and long durations are lower than that of the short durations. This is consistent to the gap between class-balanced metrics and average accuracy. BiGAT improves the average accuracies for medium and long durations by $6\% - 9\%$, compared with GCN and GAT, suggesting that incorporating spatial heterogeneity via Bimodal benefits model on challenging classes.
\begin{table}[!h]
\caption{Classification accuracies of short, medium and long outages}

\label{ClassAcc}
\begin{center}
\begin{small}
\vskip -0.15in
\begin{sc}
\begin{tabular}{lccc}
\toprule
Models & Short & Medium & Long \\
\midrule
GCN                 & 0.94         & 0.75     & 0.65 \\
GAT                 & 0.93          & 0.76     & 0.64 \\
BiGAT               & \textbf{0.95}         & 0.80     & \textbf{0.71} \\
BiGATud   & \textbf{0.95}          & \textbf{0.84}     & 0.67 \\
\midrule
\multicolumn{4}{p{0.3\textwidth}}{\tiny \textit{Note: Bold numbers indicate the best performance for a given class. BiGATud denotes BiGAT undirected.}}
\end{tabular}
\end{sc}
\end{small}
\end{center}
\vskip -0.05in
\end{table}

\section{Conclusion}

We present a machine learning framework to enhance energy infrastructure and community resilience under natural disasters. We have developed a graph-based neural network, BiGAT in particular, that predicts power outage durations using data from four Atlantic land-falling hurricanes in the Southeastern U.S. The trained BiGAT outperforms the state-of-the-art models by $2\%-15\%$. By accounting the spatial heterogeneity, BiGAT further improves the performance on medium and long duration outages.

This work has a significant real-world impact. To our knowledge, this is one of the first works that develop GNNs to learn from real data on severe weather-induced outage durations. The performance makes GNN a strong candidate for application: recovery estimation from GNN would help emergency responders and policy makers prioritize resource allocations, reduce the impact to customers, and in extreme cases save lives. This is needed for all communities, particularly socioeconomically vulnerable ones.

Our study lays the foundation for future research. Tested models show lower performance for medium and long outage classes, highlighting the need to analyze limitations with biased dataset. Future work could incorporate model-based analytical studies, such as feature importance analysis and counterfactual tests, to understand the drivers of outage duration. In this study, BiGAT was trained and evaluated on individual hurricane events in the Southeastern U.S. Generalization to unseen events or other disaster contexts remains an open challenge. Nevertheless, as a generic framework for modeling spatial characteristics, BiGAT can be extended to new hazards and regions with appropriate data and adjustments to input variables. Future studies may extend BiGAT to learning from multiple events jointly, enhancing its robustness and real-world value for resilient energy infrastructure and communities.


\bibliographystyle{IEEEtran}
\bibliography{BiGAT}

@Manual{doe2022playbook,
  title  = {Energy EmergencyResponse Playbookfor States and Territories},
  author = {{U.S. Department of Energy}},
  month  = may,
  year   = {2022}
}

@InProceedings{owerko2018predicting,
  author       = {Owerko, Damian and Gama, Fernando and Ribeiro, Alejandro},
  booktitle    = {2018 ieee global conference on signal and information processing (globalsip)},
  title        = {Predicting power outages using graph neural networks},
  year         = {2018},
  organization = {IEEE},
  pages        = {743--747},
}

@Article{velivckovic2017graph,
  author  = {Veli{\v{c}}kovi{\'c}, Petar and Cucurull, Guillem and Casanova, Arantxa and Romero, Adriana and Lio, Pietro and Bengio, Yoshua},
  journal = {arXiv preprint arXiv:1710.10903},
  title   = {Graph attention networks},
  year    = {2017},
}

@Manual{doe2021climate,
  title  = {Climate Adaptationand Resilience Plan},
  author = {{U.S. Department of Energy}},
  month  = aug,
  year   = {2021},
  url    = {https://www.sustainability.gov/pdfs/doe-2021-cap.pdf},
}

@Manual{doe2021j40,
  title  = {Justice40 Initiative},
  author = {{U.S. Department of Energy}},
  year   = {2021}
}

@Manual{white2021j40,
  title  = {Justice40},
  author = {{The White House}},
  year   = {2021}
}

@Article{ganz2023socioeconomic,
  author    = {Ganz, Scott C and Duan, Chenghao and Ji, Chuanyi},
  journal   = {PNAS nexus},
  title     = {Socioeconomic vulnerability and differential impact of severe weather-induced power outages},
  year      = {2023},
  number    = {10},
  pages     = {pgad295},
  volume    = {2},
  publisher = {Oxford University Press US},
}

@Article{afsharinejad2021large,
  author    = {Afsharinejad, Amir Hossein and Ji, Chuanyi and Wilcox, Robert},
  journal   = {Joule},
  title     = {Large-scale data analytics for resilient recovery services from power failures},
  year      = {2021},
  number    = {9},
  pages     = {2504--2520},
  volume    = {5},
  publisher = {Elsevier},
}

@Article{liao2021review,
  author    = {Liao, Wenlong and Bak-Jensen, Birgitte and Pillai, Jayakrishnan Radhakrishna and Wang, Yuelong and Wang, Yusen},
  journal   = {Journal of Modern Power Systems and Clean Energy},
  title     = {A review of graph neural networks and their applications in power systems},
  year      = {2021},
  number    = {2},
  pages     = {345--360},
  volume    = {10},
  publisher = {SGEPRI},
}

@article{flanagan2018measuring,
  title={Measuring community vulnerability to natural and anthropogenic hazards: the Centers for Disease Control and Prevention’s Social Vulnerability Index},
  author={Flanagan, Barry E and Hallisey, Elaine J and Adams, Erica and Lavery, Amy},
  journal={Journal of environmental health},
  volume={80},
  number={10},
  pages={34},
  year={2018},
  publisher={NIH Public Access}
}

@Article{moran1950notes,
  author    = {Moran, Patrick AP},
  journal   = {Biometrika},
  title     = {Notes on continuous stochastic phenomena},
  year      = {1950},
  number    = {1/2},
  pages     = {17--23},
  volume    = {37},
  publisher = {JSTOR},
}

@article{lloyd1982least,
  title={Least squares quantization in PCM},
  author={Lloyd, Stuart},
  journal={IEEE transactions on information theory},
  volume={28},
  number={2},
  pages={129--137},
  year={1982},
  publisher={IEEE}
}

@Misc{powerUS22,
  author       = {{PowerOutages.US}},
  title        = {Power outage records of the {Southeast} {US}},
  year         = {2021},
  comment      = {Poweroutage.us dataset retrived in 2021 at https://poweroutage.us/products}
}

@Misc{USCensusBureau2020,
    author = {{U.S. Census Bureau}},
    title= {American Community Survey 1-Year Estimates},
    year = {2020}
}

@Misc{HURDAT2,
  author       = {{National Hurricane Center}},
  title        = {The revised Atlantic hurricane database (HURDAT2)},
  year         = {2023}
}

@article{jaech2018real,
  title={Real-time prediction of the duration of distribution system outages},
  author={Jaech, Aaron and Zhang, Baosen and Ostendorf, Mari and Kirschen, Daniel S},
  journal={IEEE Transactions on Power Systems},
  volume={34},
  number={1},
  pages={773--781},
  year={2018},
  publisher={IEEE}
}

@article{carvallo2024weather,
  title={Weather-related Power Outages Rising},
  author={Carvallo, Juan Pablo and Casey, Joan},
  year={2024},
  publisher={Climate Central}
}

@article{DOE2018report,
  title={Department of Energy Report Explores U.S. Advanced Small Modular Reactors to Boost Grid Resiliency},
  author={{Office of Nuclear Energy}},
  year={2018},
  publisher={Department of Energy}
}

@article{nateghi2014forecasting,
  title={Forecasting hurricane-induced power outage durations},
  author={Nateghi, Roshanak and Guikema, Seth D and Quiring, Steven M},
  journal={Natural hazards},
  volume={74},
  pages={1795--1811},
  year={2014},
  publisher={Springer}
}

@article{fang2022failure,
  title={A failure prediction method of power distribution network based on PSO and XGBoost},
  author={Fang, Jian and Wang, Hongbin and Yang, Fan and Yin, Kuang and Lin, Xiang and Zhang, Min},
  journal={Australian journal of electrical and electronics engineering},
  volume={19},
  number={4},
  pages={371--378},
  year={2022},
  publisher={Taylor \& Francis}
}

@article{johnson2024can,
  title={Can socio-economic indicators of vulnerability help predict spatial variations in the duration and severity of power outages due to tropical cyclones?},
  author={Johnson, Paul M and Jackson, Nicole D and Baroud, Hiba and Staid, Andrea},
  journal={Environmental Research Letters},
  volume={19},
  number={4},
  pages={044048},
  year={2024},
  publisher={IOP Publishing}
}

@article{stankovic2022methods,
  title={Methods for analysis and quantification of power system resilience},
  author={Stankovi{\'c}, Aleksandar M and Tomsovic, Kevin L and De Caro, Fabrizio and Braun, Martin and Chow, Joe H and {\v{C}}ukalevski, Ninel and Dobson, Ian and Eto, Joseph and Fink, Blair and Hachmann, Christian and others},
  journal={IEEE Transactions on Power Systems},
  volume={38},
  number={5},
  pages={4774--4787},
  year={2022},
  publisher={IEEE}
}

@article{xu2024resilience,
  title={Resilience of renewable power systems under climate risks},
  author={Xu, Luo and Feng, Kairui and Lin, Ning and Perera, ATD and Poor, H Vincent and Xie, Le and Ji, Chuanyi and Sun, X Andy and Guo, Qinglai and O’Malley, Mark},
  journal={Nature Reviews Electrical Engineering},
  volume={1},
  number={1},
  pages={53--66},
  year={2024},
  publisher={Nature Publishing Group UK London}
}

@Misc{excutiveorder2013,
    author = {{The White House}},
    title= {Transforming our Nation's Electric Grid Through Improved Siting, Permitting, and Review},
    year = {2013},
}

@article{vapnik2006transductive,
  title={Transductive inference and semi-supervised learning},
  author={Vapnik, Vladimir},
  year={2006}
}

@article{yu2018deepsolar,
  title={DeepSolar: A machine learning framework to efficiently construct a solar deployment database in the United States},
  author={Yu, Jiafan and Wang, Zhecheng and Majumdar, Arun and Rajagopal, Ram},
  journal={Joule},
  volume={2},
  number={12},
  pages={2605--2617},
  year={2018},
  publisher={Elsevier}
}

\end{document}